\documentclass[journal]{IEEEtai}

\usepackage[colorlinks,urlcolor=blue,linkcolor=blue,citecolor=blue]{hyperref}

\usepackage{color,array}

\usepackage{graphicx}
\usepackage{tikz}
\usepackage{forest}
\usepackage{booktabs}
\usepackage{tabularx}
\newcolumntype{Y}{>{\raggedright\arraybackslash}X}
\usepackage{amssymb}
\usepackage{amsmath}
\usepackage{booktabs}
\usepackage{array}
\usepackage{threeparttable}
\usepackage{makecell}

\begin{document}

\title{Multimodal Large Language Model-Enabled Video Translation: A Role-Oriented Survey}

\author{
Bingzheng Qu, Kehai Chen, Xuefeng Bai, and Min Zhang%
\thanks{Bingzheng Qu, Kehai Chen, Xuefeng Bai, and Min Zhang are with the School of Computer Science and Technology, Harbin Institute of Technology (Shenzhen), Shenzhen 518055, Guangdong, China.}%
\thanks{Corresponding author: Kehai Chen (email: chenkehai@hit.edu.cn).}%
\thanks{Bingzheng Qu: qbzz@stu.hit.edu.cn; Xuefeng Bai: baixuefeng@hit.edu.cn; Min Zhang: zhangmin2021@hit.edu.cn.}%
}

\maketitle

\begin{abstract}
Recent progress in multimodal large language models (MLLMs) is reshaping video translation from a cascaded pipeline of automatic speech recognition, machine translation, text-to-speech, and lip synchronization into a unified multimodal reasoning and generation problem. High-quality video translation requires not only semantic fidelity, but also temporal alignment, speaker consistency, and emotional expressiveness across visual, acoustic, and linguistic streams. This survey provides a focused review of MLLM-enabled video translation through a role-oriented taxonomy. We organize MLLM-enabled and MLLM-relevant studies into three functional roles: Semantic Reasoner, which grounds translation in video understanding, temporal reasoning, and multimodal fusion; Expressive Performer, which supports controllable and context-aware speech generation; and Visual Synthesizer, which enables lip synchronization and visually coherent speaker rendering. We further summarize representative datasets, benchmarks, and metrics for each role, and discuss how current evaluation protocols fall short of end-to-end video translation requirements. Finally, we identify open challenges in long-form video understanding, temporal modeling, multimodal alignment, multilingual robustness, and responsible deployment, outlining future directions for natural and trustworthy cross-lingual video communication.
\end{abstract}

\begin{IEEEImpStatement}
Video translation is increasingly important for multilingual communication in education, entertainment, conferencing, and public services. Yet existing systems often handle speech recognition, machine translation, speech synthesis, and lip synchronization as separate modules, weakening semantic fidelity, speaker consistency, timing, and emotional expressiveness. This survey shows how multimodal large language models can support more integrated video translation by coordinating visual, acoustic, and textual reasoning. Through a three-role taxonomy of Semantic Reasoner, Expressive Performer, and Visual Synthesizer, it connects advances in video understanding, controllable speech generation, and visual synthesis. It also organizes datasets, benchmarks, and metrics, helping guide more natural, accessible, and trustworthy cross-lingual video translation systems.
\end{IEEEImpStatement}


\section{Introduction}
\label{sec:introduction}

\IEEEPARstart{V}{IDEO} translation aims to make video content accessible across languages while preserving the semantic, acoustic, and visual experience of the original content. Unlike text-only machine translation or speech translation, video translation requires a system to jointly maintain semantic fidelity, speaker identity, temporal alignment, lip synchronization, and emotional expressiveness. These requirements are increasingly important in multilingual video conferencing, online education, entertainment, social media, and public communication, where translated videos must remain both linguistically accurate and perceptually natural.

Conventional video translation systems are usually built as cascaded pipelines involving automatic speech recognition (ASR), machine translation (MT), text-to-speech (TTS), and lip synchronization~\cite{hinton2012deep,vaswani2017attention,wang2017tacotron,prajwal2020lip,jia2019direct}. Although such pipelines are modular and practical, they often suffer from error propagation and weak cross-modal coordination. For example, ASR errors can distort downstream translation, translated speech may fail to match the timing and emotion of the source video, and lip-sync modules may generate visually plausible but semantically disconnected facial motions. These limitations suggest that video translation should be studied not only as a sequence of independent subtasks, but also as an integrated multimodal reasoning and generation problem.

Recent advances in multimodal large language models (MLLMs) provide a new foundation for this integrated perspective. MLLMs have advanced from vision-language pretraining and multimodal instruction tuning~\cite{radford2021learning,alayrac2022flamingo,li2023blip,liu2023visual,ye2024mplug} to video-centric understanding, temporal reasoning, and audio-visual modeling~\cite{maaz2024video,zhang2023video,li2024llama,wang2024internvideo2}. Meanwhile, large-scale speech generation and video generation models have improved controllability, expressiveness, and visual realism~\cite{Du2024CosyVoice2,chen2025f5,xing2024survey,peng2025omnisync}. These advances make it possible to reconsider video translation as a unified process in which visual, acoustic, and textual signals are jointly interpreted, translated, synthesized, and aligned. However, existing studies are still scattered across video understanding, speech synthesis, speech translation, talking-head generation, and video generation, leaving a lack of systematic organization for MLLM-enabled video translation. In this survey, we use the term MLLM-enabled video translation to denote systems and methods in which multimodal language models provide semantic understanding, cross-modal reasoning, planning, or coordination for translating videos. We distinguish these core MLLM components from supporting generation modules, such as expressive TTS, lip synchronization, talking-head animation, and enabling visual backbones. The latter are included only when their mechanisms are directly relevant to translated-video generation, such as speaker preservation, duration control, emotion transfer, audio-visual synchronization, or identity-consistent rendering.

\begin{table*}[t]
\centering
\caption{Comparison with existing surveys related to MLLM-enabled video translation.}
\label{tab:related_surveys}
\footnotesize
\renewcommand{\arraystretch}{1.12}
\setlength{\tabcolsep}{4pt}
\begin{threeparttable}
\begin{tabularx}{0.95\textwidth}{@{}>{\raggedright\arraybackslash}p{0.14\textwidth}
                              >{\centering\arraybackslash}p{0.040\textwidth}
                              >{\raggedright\arraybackslash}p{0.21\textwidth}
                              >{\centering\arraybackslash}p{0.03\textwidth}
                              >{\centering\arraybackslash}p{0.03\textwidth}
                              >{\centering\arraybackslash}p{0.03\textwidth}
                              >{\raggedright\arraybackslash}X@{}}
\toprule
\textbf{Survey} & \textbf{Year} & \textbf{Main Scope} &
\textbf{SR} & \textbf{EP} & \textbf{VS} &
\textbf{Missing Link} \\
\midrule
Shen et al.~\cite{shen2024survey}
& 2024 & Multimodal MT
& $\triangle$ & -- & --
& Multimodal translation without speech or visual synthesis \\

Das et al.~\cite{das2025video}
& 2025 & Video-guided MT
& $\checkmark$ & -- & --
& Video-guided text translation only \\

Yin et al.~\cite{yin2024survey}
& 2024 & General MLLMs
& $\checkmark$ & $\triangle$ & $\triangle$
& Broad MLLM scope; not video-translation-specific \\

Tang et al.~\cite{tang2025video}
& 2025 & Video Understanding
& $\checkmark$ & -- & --
& Video comprehension without cross-lingual generation \\

Wu et al.~\cite{wu2025survey}
& 2025 & Temporal Grounding
& $\checkmark$ & -- & --
& Temporal localization without translation/generation \\

Gupta et al.~\cite{gupta2024direct}
& 2024 & Speech-to-speech Translation
& $\triangle$ & $\checkmark$ & --
& Speech translation with limited visual synchronization \\

Xie et al.~\cite{xie2025towards}
& 2025 & Controllable TTS
& -- & $\checkmark$ & --
& Speech controllability without video grounding \\

Xing et al.~\cite{xing2024survey}
& 2024 & Video Diffusion
& -- & -- & $\triangle$
& Generic generation without translation constraints \\

Rakesh et al.~\cite{rakesh2025advancing}
& 2025 & Talking-head Generation
& -- & $\triangle$ & $\checkmark$
& Speaker animation without semantic translation \\
\bottomrule
\end{tabularx}
\begin{tablenotes}
\footnotesize
\item SR: Semantic Reasoner; EP: Expressive Performer; VS: Visual Synthesizer. $\checkmark$ indicates central coverage, $\triangle$ indicates partial or related coverage.
\end{tablenotes}
\vspace{-3mm}
\end{threeparttable}
\end{table*}

To address this gap, this survey provides a role-oriented review of MLLM-enabled video translation. Instead of treating ASR, MT, TTS, and lip synchronization as isolated modules, we organize the literature according to the functional roles that MLLMs and MLLM-related models can play in the translation pipeline. The main contributions are summarized as follows:

\begin{itemize}

\item \textbf{Role-oriented framework for MLLM-enabled video translation:}
We formulate video translation as an integrated multimodal reasoning and generation problem, where visual, acoustic, and textual signals are jointly used to preserve semantic fidelity, temporal alignment, speaker consistency, and emotional expressiveness.

\item \textbf{Three-role taxonomy of existing methods:}
We organize related studies into three functional roles: \emph{Semantic Reasoner}, which supports video understanding, temporal reasoning, and multimodal grounding; \emph{Expressive Performer}, which supports controllable and context-aware speech generation; and \emph{Visual Synthesizer}, which supports lip synchronization and visually coherent speaker rendering.

\item \textbf{Role-oriented analysis of datasets, benchmarks, and metrics:}
We summarize representative resources and evaluation protocols for each role, and discuss the gap between current proxy evaluations and the requirements of end-to-end video translation.

\item \textbf{Discussion of open challenges and future directions:}
We identify key challenges in long-form video understanding, temporal modeling, multimodal alignment, multilingual robustness, real-time deployment, and responsible use, providing a roadmap for future MLLM-enabled video translation systems.

\end{itemize}

The remainder of this survey is organized as follows. Section~\ref{sec:survey_methodology_and_scope} describes the literature search protocol and the relation to existing surveys. Section~\ref{sec:preliminaries} introduces the preliminaries of video translation and MLLMs. Section~\ref{sec:taxonomy} presents the proposed role-oriented taxonomy. Section~\ref{sec:evaluation} reviews representative benchmarks, datasets, and evaluation metrics. Section~\ref{sec:limitations} discusses limitations and future research directions. Section~\ref{sec:conclusion} concludes the survey.

\section{Survey Methodology and Scope}
\label{sec:survey_methodology_and_scope}
\subsection{Literature Search Protocol}
To ensure focused coverage, we adopted a role-oriented literature search protocol. We searched IEEE Xplore, ACM Digital Library, ACL Anthology, arXiv, Google Scholar, and Semantic Scholar for studies published up to May 2026. We also traced references from representative and highly cited studies to include relevant foundational works.
Studies were included if they contributed to at least one functional role in MLLM-enabled video translation: semantic reasoning over video, expressive speech generation, visual synchronization or speaker rendering, or benchmark and evaluation design. We excluded generic image-language, TTS, or video generation works unless they provide mechanisms directly relevant to video translation, such as multimodal grounding, cross-lingual transfer, expressive speech rendering, duration control, lip synchronization, identity preservation, or temporally coherent speaker synthesis. For duplicate or extended versions, we prioritized the peer-reviewed, latest, or most complete version. The retained studies were organized according to their functional role, modality coverage, task formulation, and evaluation protocol.

\subsection{Relation to Existing Surveys}
Existing surveys have covered related areas, including multimodal and video-guided machine translation, general MLLMs, video understanding with LLMs, speech-to-speech translation, controllable speech synthesis, video diffusion, and talking-head generation~\cite{shen2024survey,das2025video,yin2024survey,tang2025video,gupta2024direct,xie2025towards,xing2024survey,rakesh2025advancing}. However, they typically address isolated components rather than the full path from multimodal grounding to target-language speech and visual rendering. In contrast, this survey treats MLLM-enabled video translation as an integrated reasoning-and-generation task. As summarized in Table~\ref{tab:related_surveys}, we organize existing studies into three roles: Semantic Reasoner, Expressive Performer, and Visual Synthesizer, thereby linking semantic grounding, cross-lingual transfer, expressive speech generation, and temporally aligned visual synthesis.

\section{Preliminaries}
\label{sec:preliminaries}

\subsection{Video Translation}
Video translation aims to transform a source-language video into a target-language video or subtitle stream while preserving the meaning, timing, speaker identity, prosody, emotion, and visual coherence of the original content. A conventional video translation pipeline typically consists of automatic speech recognition (ASR)~\cite{hinton2012deep}, machine translation (MT)~\cite{vaswani2017attention}, text-to-speech (TTS)~\cite{wang2017tacotron}, and lip synchronization~\cite{prajwal2020lip}. Its component technologies have evolved from classical and neural ASR~\cite{rabiner1993fundamentals,graves2006connectionist,hannun2014deep,chan2015listen}, statistical and neural MT~\cite{brown1993mathematics,koehn2003statistical,sutskever2014sequence}, parametric and neural TTS~\cite{tokuda2000speech,zen2009statistical,van2016wavenet}, to lip synchronization and talking-head generation~\cite{cosatto2002photo,son2017lip,zakharov2019few}. Although these modules can be developed independently, their errors and design choices are tightly coupled in video translation.

ASR converts source speech into transcripts and timestamps, providing the textual and temporal basis for downstream translation. Recognition or segmentation errors may propagate to MT and alter the intended meaning or speaker attribution. MT converts source-language transcripts into target-language text, but in video translation it must also consider visual context, speaker turns, timing constraints, and cross-lingual duration differences. TTS renders the translated text into speech and therefore determines naturalness, speaker similarity, prosody, emotion, and speaking rate. Finally, lip synchronization or talking-head generation aligns the translated speech with mouth movements and facial dynamics, making upstream errors visually apparent. 

\subsection{Multimodal Large Language Models}
Multimodal large language models (MLLMs) extend large language models with visual, acoustic, or other modality inputs, enabling them to perform cross-modal understanding, reasoning, and generation~\cite{alayrac2022flamingo,liu2023visual,han2024onellm}. In typical MLLM architectures, modality-specific encoders extract visual or audio features, and projection modules or adapters map these features into token representations that can interact with the language model. Through multimodal pretraining, instruction tuning, and cross-modal alignment, MLLMs can connect language with objects, events, speakers, temporal relations, and acoustic cues. The development of MLLMs is closely related to scalable Transformer architectures, vision Transformers, contrastive vision-language pretraining, and video-language representation learning~\cite{dosovitskiy2020image,radford2021learning,jia2021scaling,sun2019videobert,zhu2020actbert,zellers2021merlot,bain2021frozen,fu2021violet,wang2022omnivl,wang2022internvideo}. More recent unified multimodal systems further extend this line of work toward image, video, audio, language, and action modeling~\cite{shukor2023unival,lu2024unified,chen2023pali,bai2023qwenvlversatilevisionlanguagemodel}.

For video translation, MLLMs are valuable not simply as generic video understanding models, but as potential coordinators of multimodal reasoning and generation. They can ground ambiguous utterances in visual context, maintain cross-shot semantic consistency, guide expressive speech synthesis, and support temporally aligned visual rendering. This motivates a role-oriented view of MLLM-enabled video translation, where different models and techniques are analyzed according to their functions in semantic reasoning, expressive speech generation, and visual synthesis.

\tikzstyle{my-box}=[ 
    rectangle, draw=gray!50, rounded corners, text opacity=1,
    minimum height=2em, minimum width=5em,
    inner sep=3pt, inner ysep=6pt, align=center,
    fill opacity=0.15, line width=0.5pt,
]
\tikzstyle{leaf}=[my-box, minimum height=2em, fill=gray!5, text=black,
    align=left,inner xsep=3pt, inner ysep=6pt, line width=0.5pt,
]

\definecolor{cCon}{RGB}{255,153,102}   % Constitution (orange)
\definecolor{cEvo}{RGB}{120,200,140}   % Evolution & Learning (green)
\definecolor{cApp}{RGB}{ 80,160,220}   % Applications (blue)
\definecolor{cEva}{RGB}{160,140,220}   % Datasets & Evaluation (purple)

\newcommand{\leafw}{27.5em}

\begin{figure*}[t!]
  \centering
  \resizebox{0.9\textwidth}{!}{
  \begin{forest}
    for tree={
      grow=east,
      reversed=true,
      anchor=base west,
      parent anchor=east,
      child anchor=west,
      base=center,
      font=\large,
      rectangle,
      rounded corners,
      draw=gray!70,
      align=left,
      text centered,
      minimum width=4em,
      edge={darkgray, line width=0.3mm},
      edge path={
        \noexpand\path [\forestoption{edge}]
          (!u.east) -- +(.8em,0) |- (.child anchor)\forestoption{edge label};
      },
      s sep=14pt,
      l sep=14pt,
      inner xsep=4pt,
      inner ysep=6pt,
      line width=0.4pt,
      ver/.style={rotate=90, child anchor=north, parent anchor=south, anchor=center},
    },
    where level=1{
      font=\normalsize, 
      text width=10em,
      edge path={
        \noexpand\path [\forestoption{edge}]
          (!u.south)   
          -- +(.8em,0) 
          |- ($(.child anchor)+(-.8em,0)$)
          -- (.child anchor)\forestoption{edge label};
      }
    }{} ,
    where level=2{font=\normalsize, text width=13em}{} ,
    where level=3{font=\normalsize, text width=12em}{} ,
    where level=4{font=\footnotesize, text width=30em, align=left, text badly centered}{},
    % ------------------------ ROOT ------------------------
    [\Large \textbf{Taxonomy}, ver, line width=0.35mm
      % ================= I. The Semantic Reasoner ==================
      [\large \shortstack{\textbf{The Semantic}\\\textbf{Reasoner}},
        fill=cApp!30, draw=cApp!80, line width=0.3mm
        [\textbf{Multimodal Grounding}, fill=cApp!18, draw=cApp!80, edge={cApp!80}
        [{Video-ChatGPT~\cite{maaz2024video}, Video-LLaMA 3~\cite{zhang2025videollama3frontiermultimodal},
            VideoChat~\cite{li2023videochat},\\
            Valley~\cite{luo2023valley}, VideoChat2~\cite{li2024mvbench},
            InternVideo2~\cite{wang2024internvideo2},OneLLM~\cite{han2024onellm},\\
            Video-LLaMA~\cite{zhang2023video}, GroundingGPT~\cite{li2024groundinggpt}, LLaMA-VID~\cite{li2024llama},},   
            leaf,text width=\leafw, draw=cApp!80, line width=0.35mm, edge={cApp!80}]
        ]
        [\textbf{Temporal \& Event Reasoning}, fill=cApp!18, draw=cApp!80, edge={cApp!80}
         [{MovieChat~\cite{song2024moviechat}, MovieLLM~\cite{song2024moviellm},
            LongVLM~\cite{weng2024longvlm}, TimeChat~\cite{ren2024timechat},\\
            VTimeLLM~\cite{huang2024vtimellm}, Momentor~\cite{qian2024momentor},
            VideoLLM-online~\cite{chen2024videollm},\\
            LITA~\cite{huang2024lita}, VTG-LLM~\cite{guo2025vtg},
            Video-XL~\cite{shu2025video}, Time-R1~\cite{wang2025time}},
            leaf,text width=\leafw, draw=cApp!80, line width=0.35mm, edge={cApp!80}]
        ]
        [\shortstack{\textbf{Cross-lingual Planning}}, fill=cApp!18, draw=cApp!80, edge={cApp!80}
          [{Direct S2ST~\cite{jia2019direct}, SEAMLESSM4T~\cite{communication2023seamlessm4t},
            AV2AV~\cite{choi2024av2av},\\
            Adaptive Inner Speech-Text Alignment~\cite{liu2025adaptive},
            InImageTrans~\cite{zuo2025inimagetrans}},
            leaf,text width=\leafw, draw=cApp!80, line width=0.35mm, edge={cApp!80}]
        ]
        [\textbf{Efficient Adaptation}, fill=cApp!18, draw=cApp!80, edge={cApp!80}
          [{LLaMA-Adapter~\cite{zhang2023llama}, BT-Adapter~\cite{liu2024bt},
            Otter~\cite{li2025otter},\\
            PLLaVA~\cite{xu2024pllava}, RED-VILLM~\cite{huang2024image},
            REEF~\cite{reza2025reef}, VLog~\cite{lin2025vlog}},
             leaf,text width=\leafw, draw=cApp!80, line width=0.35mm, edge={cApp!80}]
        ]
      ]
      % ===================== II. The Expressive Performer =====================
      [\large \shortstack{\textbf{The Expressive}\\\textbf{Performer}},
        fill=cEvo!30, draw=cEvo!80, line width=0.3mm
        [\textbf{Speech-token LM}, fill=cEvo!18, draw=cEvo!80, edge={cEvo!80}
          [{VALL-E R~\cite{han2024vall}, CosyVoice~\cite{FunAudioLLMCosyVoice},
            CosyVoice 2~\cite{Du2024CosyVoice2}, \\
            Fish-Speech~\cite{Liao2024FishSpeech}, HALL-E~\cite{Nishimura2024HALLE},
            VoxInstruct~\cite{zhou2024voxinstruct}, Spark-TTS~\cite{Wang2025SparkTTS},\\
            MegaTTS 2~\cite{jiang2024megatts2boostingprompting},
            MegaTTS 3~\cite{jiang2025megatts}, Seed-TTS~\cite{Anastassiou2024SeedTTS}},
            leaf,text width=\leafw, draw=cEvo!80, line width=0.35mm, edge={cEvo!80}]
        ]
        [\textbf{Prompt \& Emotion Control}, fill=cEvo!18, draw=cEvo!80, edge={cEvo!80}
          [{PromptTTS~\cite{guo2023prompttts}, PromptTTS 2~\cite{leng2023prompttts},
            InstructTTS~\cite{Yang2023InstructTTS},\\
            StyleTTS 2~\cite{Li2023StyleTTS2}, ControlSpeech~\cite{ji2025controlspeech},
            SC VALL-E~\cite{kim2023sc}},
            leaf,text width=\leafw, draw=cEvo!80, line width=0.35mm, edge={cEvo!80}]
        ]
        [\textbf{Diffusion \& Flow Rendering}, fill=cEvo!18, draw=cEvo!80, edge={cEvo!80}
          [{F5-TTS~\cite{chen2025f5}, E2 TTS~\cite{Eskimez2024E2TTS},
            Voicebox~\cite{le2023voicebox}, NaturalSpeech 2~\cite{Shen2023NaturalSpeech2},\\
            NaturalSpeech 3~\cite{Ju2024NaturalSpeech3}, DiTTO-TTS~\cite{Lee2024DiTToTTS},
            Mask-GCT~\cite{wang2024maskgct},\\
            SimpleSpeech 2~\cite{yang2025simplespeech2}, FlashSpeech~\cite{ye2024flashspeech},
            DEX-TTS~\cite{Park2024DEXTTS}},
            leaf,text width=\leafw, draw=cEvo!80, line width=0.35mm, edge={cEvo!80}]
        ]
      ]
      % ================ III. The Visual Synthesizer=================
      [\large \shortstack{\textbf{The Visual}\\\textbf{Synthesizer}},
        fill=cEva!30, draw=cEva!80, line width=0.3mm
        [\shortstack{\textbf{Lip Synchronization}}, fill=cEva!18, draw=cEva!80, edge={cEva!80}
          [{Wav2Lip~\cite{prajwal2020lip}, TalkLip~\cite{zhou2023seeing},
            VideoReTalking~\cite{cheng2022videoretalking}, Diff2Lip~\cite{mukhopadhyay2024diff2lip},\\
            MuseTalk~\cite{zhang2024musetalk}, LatentSync~\cite{li2024latentsync},
            OmniSync~\cite{peng2025omnisync}},
            leaf,text width=\leafw, draw=cEva!80, line width=0.35mm, edge={cEva!80}]
        ]
        [\shortstack{\textbf{Talking-head Animation}}, fill=cEva!18, draw=cEva!80, edge={cEva!80}
          [{Few-shot Talking Head~\cite{zakharov2019few}, IP-LAP~\cite{nie2023identity},
            Kling-Avatar~\cite{ding2025kling},\\
            OmniHuman-1~\cite{lin2025omnihuman1}, MIDAS~\cite{chen2025midas},
            Empathyear~\cite{fei2024empathyear}},
            leaf,text width=\leafw, draw=cEva!80, line width=0.35mm, edge={cEva!80}]
        ]
        [\textbf{Enabling Visual Backbones}, fill=cEva!18, draw=cEva!80, edge={cEva!80}
          [{Stable Video Diffusion~\cite{blattmann2023stable},
            AnimateDiff~\cite{guo2023animatediff}, \\
            Tune-A-Video~\cite{wu2023tune}, CogVideoX~\cite{yang2024cogvideox},
            VideoCrafter~\cite{chen2023videocrafter1},\\
            Phantom~\cite{liu2025phantom}, HunyuanVideo~\cite{kong2024hunyuanvideo},
            Wan~\cite{wan2025wan}, Vidu~\cite{bao2024vidu}},
            leaf,text width=\leafw, draw=cEva!80, line width=0.35mm, edge={cEva!80}]
        ]
      ]
    ] % end ROOT
  \end{forest}
  }% end
  \caption{Role-oriented taxonomy of MLLM-enabled video translation. The taxonomy separates three levels of relevance: core MLLM reasoning methods, translation-supporting speech and visual generation methods, and enabling generative backbones. Core methods directly perform multimodal understanding or cross-lingual planning, while supporting and enabling methods are discussed only through their contribution to translated-video quality.}
  \label{fig:taxonomy}
\end{figure*}

\section{A Multi-Role Taxonomy for MLLM-Enabled Video Translation}
\label{sec:taxonomy}

Building on the role-oriented framework introduced above, we propose a multi-role taxonomy for MLLM-enabled video translation. As shown in Fig.~\ref{fig:taxonomy}, the taxonomy organizes existing methods according to their functional contribution to the translation pipeline rather than by isolated task boundaries such as ASR, MT, TTS, and lip synchronization. Specifically, we divide MLLM-enabled video translation into three interconnected roles: the \emph{Semantic Reasoner}, which interprets source videos and produces translation-aware multimodal representations; the \emph{Expressive Performer}, which renders translated content into natural and contextually appropriate speech; and the \emph{Visual Synthesizer}, which aligns translated speech with visually coherent speaker motion. This role-oriented view clarifies how recent advances in video understanding, speech generation, and visual synthesis can jointly support high-quality cross-lingual video communication.

\begin{itemize}
  \item \textbf{The Semantic Reasoner (Section~\ref{sec:semantic_reasoner})}: 
  This role focuses on understanding the source video before and during translation. It grounds linguistic content in visual, acoustic, and temporal context, including speaker attribution, event boundaries, visual references, emotional cues, and long-range temporal structure. In our taxonomy, this role includes multimodal grounding, temporal and event reasoning, cross-lingual planning, and efficient adaptation.

  \item \textbf{The Expressive Performer (Section~\ref{sec:expressive_performer})}: 
  This role focuses on generating target-language speech that is not only intelligible and semantically faithful, but also expressive and synchronized with the source video. It covers speech-token language modeling, prompt- and emotion-controlled TTS, and diffusion- or flow-based acoustic rendering. These techniques support speaker similarity, prosody, rhythm, emotion, and duration control in translated speech.

  \item \textbf{The Visual Synthesizer (Section~\ref{sec:visual_synthesizer})}: 
  This role focuses on rendering the translated speech into visually coherent speaker motion. It includes lip synchronization, talking-head animation, and enabling visual backbones that support identity preservation, mouth-shape alignment, facial expression consistency, and temporal coherence. This role reduces perceptual mismatch between synthesized speech and on-screen appearance.
\end{itemize}

\begin{figure*}[t!]\centering
  \includegraphics[width=\linewidth]{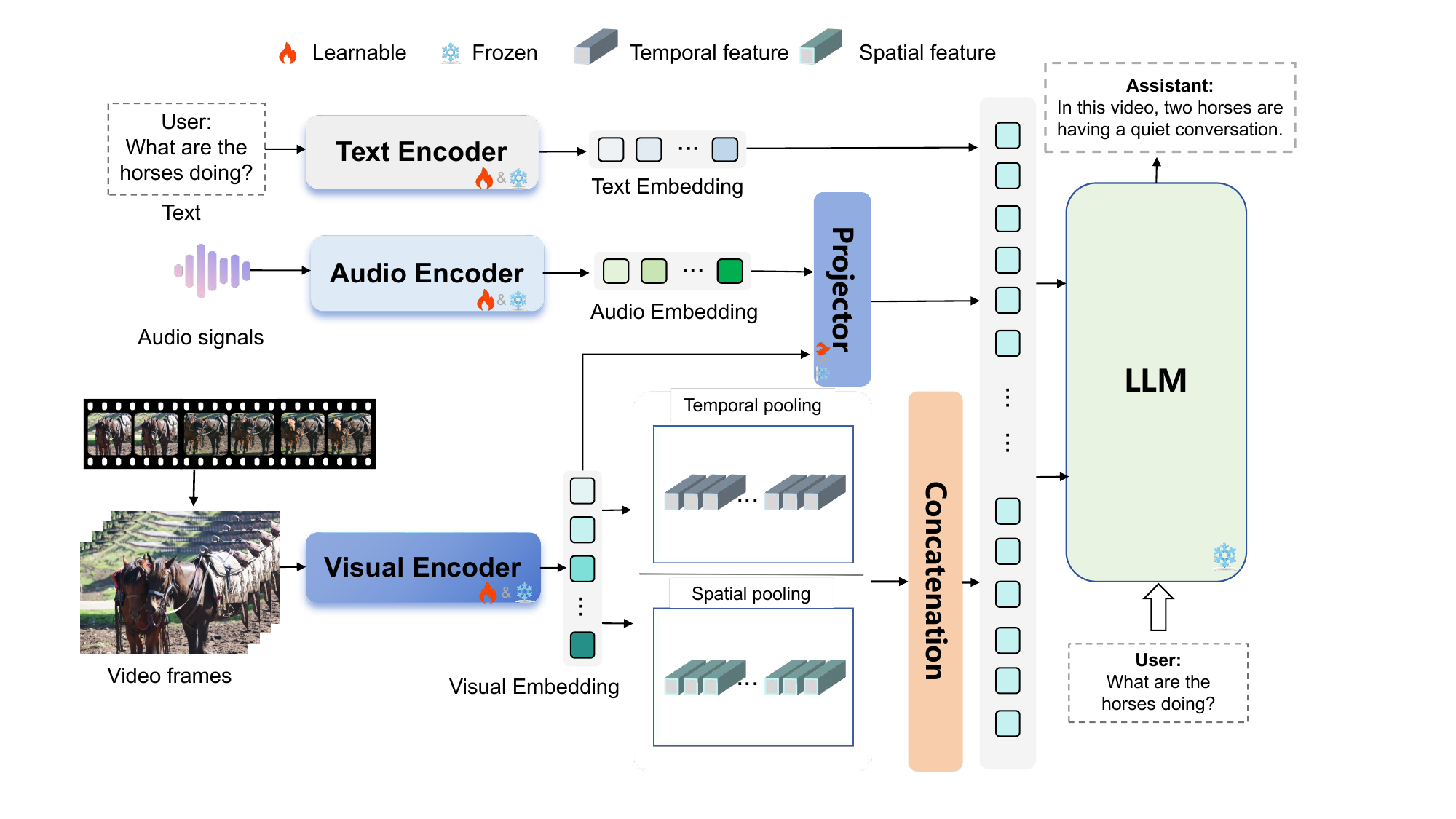}
  \caption{Typical architecture of an MLLM-enabled video understanding model. Text, audio, and video inputs are encoded by trainable or frozen modality-specific encoders. The resulting embeddings are projected, temporally and spatially pooled, and concatenated into multimodal tokens.}
\label{fig:mllm_video_understanding}
\end{figure*}

\subsection{The Semantic Reasoner}
\label{sec:semantic_reasoner}

As the comprehension and inference core of MLLM-enabled video translation, the Semantic Reasoner is responsible for converting source videos into semantically grounded, temporally organized, and translation-aware representations. Unlike generic video understanding, semantic reasoning for video translation must support cross-lingual meaning transfer while preserving speaker attribution, event boundaries, visual context, emotional cues, and temporal order. These capabilities are essential because many translation decisions cannot be reliably inferred from ASR transcripts alone, especially in videos with multiple speakers, off-screen references, scene changes, gestures, or visually grounded expressions. As shown in Fig.~\ref{fig:taxonomy}, we organize Semantic Reasoner methods into four functional categories: \emph{Multimodal Grounding}, \emph{Temporal/Event Reasoning}, \emph{Cross-lingual Planning}, and \emph{Efficient Adaptation}.

\subsubsection{Multimodal Grounding}

Multimodal grounding establishes the basic interface between video content and language reasoning. In most video-MLLM pipelines, visual frames, audio signals, or clip-level features are first encoded by modality-specific encoders and then projected into the token space of a language model through linear projection, Q-Former-style modules, cross-attention, or other adapter mechanisms. Once aligned with textual prompts, these multimodal tokens allow the LLM to perform question answering, instruction following, and semantic interpretation over video content.

Representative methods differ in how they align visual, acoustic, and textual information. Video-ChatGPT~\cite{maaz2024video} adapts visual encoders for video instruction following and introduces video-oriented instruction data and evaluation. Video-LLaMA~\cite{zhang2023video} incorporates both visual and audio branches, making it more relevant to translation scenarios where acoustic events and speech cues complement visual context. LLaMA-VID~\cite{li2024llama} improves efficiency by representing each frame with compact context and content tokens, while VideoChat~\cite{li2023videochat}, Valley~\cite{luo2023valley}, VideoChat2~\cite{li2024mvbench}, InternVideo2~\cite{wang2024internvideo2}, and Video-LLaMA 3~\cite{zhang2025videollama3frontiermultimodal} further strengthen video-language alignment through improved video encoders or large-scale multimodal pretraining. For video translation, such grounding is crucial for resolving visually dependent ambiguities, identifying speakers and objects, and maintaining consistency between translated expressions and the visual scene.

\subsubsection{Temporal \& Event Reasoning}

Beyond static grounding, video translation requires temporal and event-level reasoning. A translation system must determine when an utterance occurs, which visual event it refers to, how speaker turns evolve, and how long the translated output can last without breaking audiovisual synchronization. This makes long-context modeling, temporal localization, event segmentation, and motion-aware reasoning central to the Semantic Reasoner.

Recent video-MLLMs address these challenges from different perspectives. MovieChat~\cite{song2024moviechat}, MovieLLM~\cite{song2024moviellm}, LongVLM~\cite{weng2024longvlm}, MA-LMM~\cite{he2024ma}, and Video-XL~\cite{shu2025video} focus on long-form video comprehension by introducing memory mechanisms, hierarchical representations, or efficient long-video processing. TimeChat~\cite{ren2024timechat}, VTimeLLM~\cite{huang2024vtimellm}, Momentor~\cite{qian2024momentor}, LITA~\cite{huang2024lita}, VTG-LLM~\cite{guo2025vtg}, and Time-R1~\cite{wang2025time} emphasize temporal grounding, moment localization, and fine-grained event reasoning. VideoLLM-online~\cite{chen2024videollm} further explores streaming video understanding, which is important for real-time translation settings. These methods help video translation systems preserve temporal coherence, avoid mistranslating visually grounded events, and align translated speech with actions, pauses, and scene transitions. Related dense captioning, long-form pretraining, streaming understanding, and temporal search studies further show that event abstraction, temporal retrieval, and keyframe selection are important for scaling video reasoning to realistic translation scenarios~\cite{zhang2024simple,yang2024vript,ye2025re,cheng2025scaling,tang2025adaptive,upadhyay2025time}.

\subsubsection{Cross-lingual Planning}

Cross-lingual planning connects multimodal understanding with translation-oriented decision making. While many MLLMs can describe or answer questions about videos, video translation additionally requires planning how source-language meaning should be transferred into target-language text or speech under timing, speaker, and modality constraints. This category therefore includes methods that support speech-to-speech translation, multimodal translation, or visually grounded machine translation.

Early direct speech-to-speech translation systems~\cite{jia2019direct} reduce error propagation by mapping source speech directly to target speech or target representations. SEAMLESSM4T~\cite{communication2023seamlessm4t} extends this direction toward massively multilingual and multimodal translation, providing a foundation for multilingual speech and text transfer. More recent audio-visual translation systems, such as AV2AV~\cite{choi2024av2av}, explore direct audio-visual speech-to-audio-visual speech translation, which is closely aligned with the goal of translated-video generation. In addition, adaptive speech-text alignment~\cite{liu2025adaptive} and multimodal text-image translation~\cite{zuo2025inimagetrans} highlight the importance of aligning linguistic transfer with non-textual context. In MLLM-enabled video translation, cross-lingual planning can guide what should be translated, when it should be spoken, which contextual cues should be preserved, and how the translation should interact with downstream speech and visual synthesis modules.

\subsubsection{Efficient Adaptation}

Efficient adaptation determines whether large multimodal models can be practically applied to video translation. Since video inputs are long, redundant, and computationally expensive, directly fine-tuning full-scale MLLMs is often infeasible. Parameter-efficient tuning, lightweight temporal modules, visual token reduction, and reuse of pretrained image-language models are therefore important for building scalable Semantic Reasoners.

LLaMA-Adapter~\cite{zhang2023llama} demonstrates that instruction tuning can be achieved by inserting lightweight learnable prompts into a frozen language model. BT-Adapter~\cite{liu2024bt} extends this idea to video by introducing a temporal branch that enriches visual representations without fully updating the backbone. Otter~\cite{li2025otter} explores multimodal in-context instruction tuning, while PG-Video-LLaVA~\cite{munasinghe2023pg}, PLLaVA~\cite{xu2024pllava}, RED-VILLM~\cite{huang2024image}, REEF~\cite{reza2025reef}, and VLog~\cite{lin2025vlog} investigate different strategies for video instruction tuning, efficient adaptation, or relevance-aware video representation. For video translation, these methods are important because practical systems must process long videos, multiple speakers, and streaming inputs under memory and latency constraints. Efficient adaptation also makes it easier to specialize a general MLLM toward translation-specific reasoning without retraining.

\subsection{The Expressive Performer}
\label{sec:expressive_performer}

The Expressive Performer is responsible for rendering translated linguistic content into natural, intelligible, and contextually appropriate speech. In video translation, speech generation is not merely a text-to-audio conversion problem. The generated speech must preserve speaker identity, prosody, rhythm, emotion, speaking style, and duration constraints so that it remains consistent with the source video and can be synchronized with the visual stream. This requirement makes expressive speech generation a central component of MLLM-enabled video translation. As shown in Fig.~\ref{fig:taxonomy}, we organize related methods into three functional categories: \emph{Speech-token Language Modeling}, \emph{Prompt \& Emotion Control}, and \emph{Diffusion \& Flow Rendering}.

\subsubsection{Speech-token Language Modeling}

\begin{figure}[htbp]\centering
  \includegraphics[width=\linewidth]{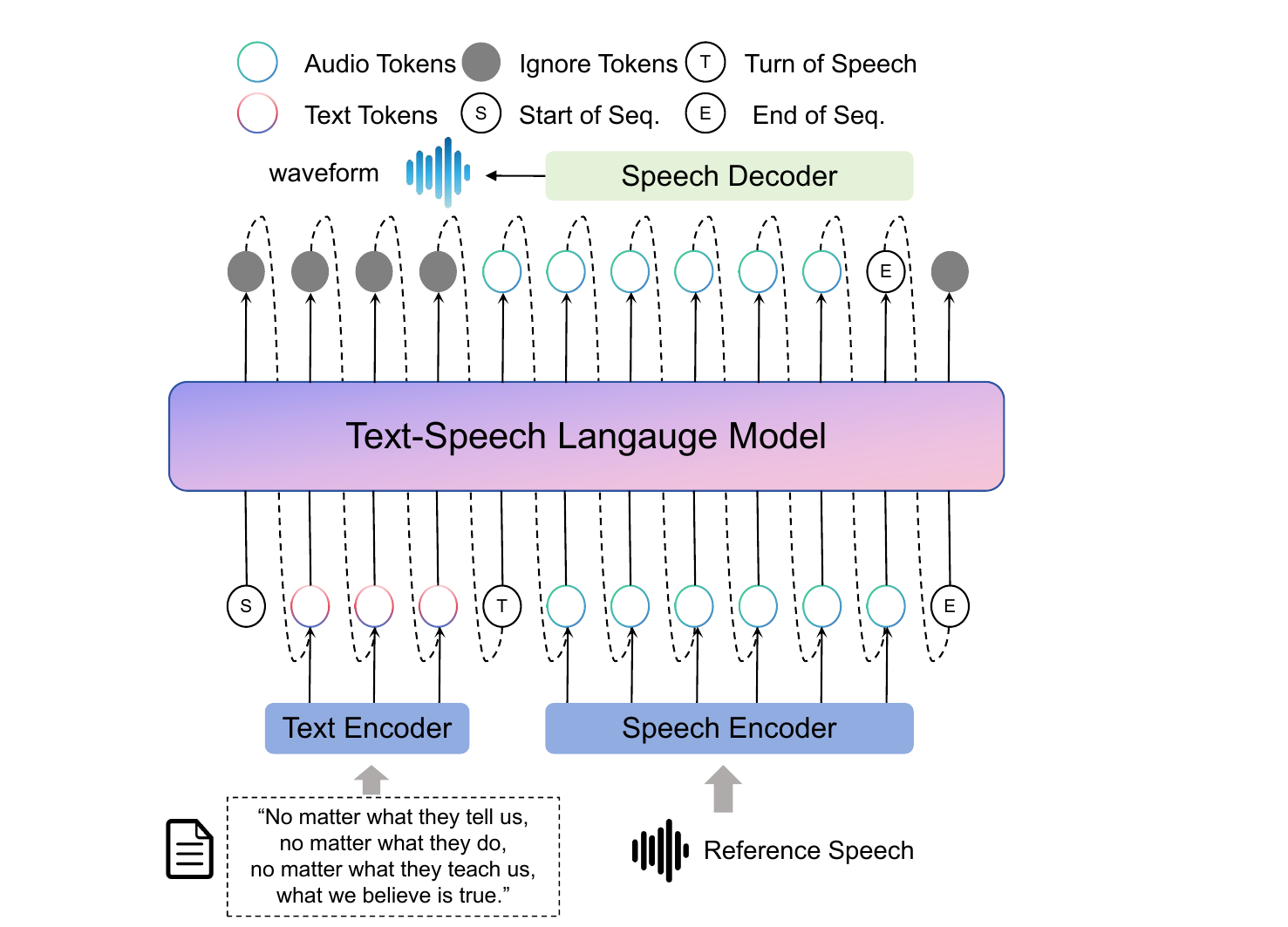}
  \caption{A typical architecture of MLLM-enabled text-to-speech synthesis.}
  \label{fig:mllm_tts_architecture}
\end{figure}

Speech-token language modeling treats speech generation as a sequence modeling problem over semantic, acoustic, or neural codec tokens. Instead of directly predicting waveforms, these methods first represent speech as discrete or structured token sequences and then use language-model-like architectures to generate target speech. This paradigm is particularly suitable for video translation because translated speech must preserve linguistic content while allowing control over speaker identity, style, and temporal structure. A typical MLLM-enabled TTS architecture is illustrated in Fig.~\ref{fig:mllm_tts_architecture}.

Recent systems demonstrate the effectiveness of this paradigm. VALL-E R~\cite{han2024vall} improves the robustness of neural codec language modeling through monotonic alignment between phoneme and acoustic token sequences. CosyVoice~\cite{FunAudioLLMCosyVoice} and CosyVoice 2~\cite{Du2024CosyVoice2} use language-model-based semantic token prediction together with flow-based acoustic modeling to support scalable multilingual and streaming speech synthesis. Spark-TTS~\cite{Wang2025SparkTTS} adopts a single-stream token design with a BiCodec representation to enable fine-grained acoustic control. Fish-Speech~\cite{Liao2024FishSpeech}, HALL-E~\cite{Nishimura2024HALLE}, VoxInstruct~\cite{zhou2024voxinstruct}, MegaTTS 2~\cite{jiang2024megatts2boostingprompting}, MegaTTS 3~\cite{jiang2025megatts}, and Seed-TTS~\cite{Anastassiou2024SeedTTS} further explore codec language modeling, hierarchical token generation, prompting mechanisms, and large-scale speech generation. For video translation, these methods provide a flexible basis for generating target-language speech while retaining speaker similarity and expressive variability.

\subsubsection{Prompt \& Emotion Control}

Prompt- and emotion-controlled TTS focuses on controlling how the translated speech should be spoken. In video translation, the target utterance should not only be semantically correct but also match the emotional tone, communicative intent, and audiovisual context of the original video. Natural-language prompts, style descriptions, reference speech, and preference signals provide convenient interfaces for controlling prosody, emotion, speaking rate, and vocal style.
Earlier controllable and style-aware TTS systems, such as StyleTTS~\cite{li2023styletts}, GenerSpeech~\cite{huang2022generspeech}, DurIAN-E~\cite{gu2023durian}, GradStyleSpeech~\cite{kang2023gradstylespeech}, and CLAM-TTS~\cite{kim2024clam}, also provide important foundations for modeling speaker style, prosody, and expressive variation.

PromptTTS~\cite{guo2023prompttts} and PromptTTS 2~\cite{leng2023prompttts} introduce text-prompt-based speech style control, enabling synthesis systems to follow natural-language descriptions of voice characteristics. InstructTTS~\cite{Yang2023InstructTTS} extends this idea by modeling expressive TTS in a discrete latent space guided by style instructions. EMO-DPO~\cite{Gao2024EmoDPO} uses direct preference optimization to improve emotional controllability, while StyleTTS 2~\cite{Li2023StyleTTS2} enhances naturalness and style modeling through diffusion and adversarial training with speech language modeling. ControlSpeech~\cite{ji2025controlspeech} targets simultaneous control of speaker cloning and language style, and SC VALL-E~\cite{kim2023sc} introduces style-controllable zero-shot speech synthesis. These methods are important because emotional mismatch, unnatural prosody, or incorrect rhythm can make a translated video perceptually inconsistent despite accurate textual translation.

\subsubsection{Diffusion \& Flow Rendering}

Diffusion- and flow-based speech rendering methods focus on high-fidelity acoustic generation. While speech-token language models provide flexible linguistic and speaker-level modeling, the final speech waveform or acoustic representation must still be rendered with high naturalness, stability, and temporal precision. Diffusion models and flow-matching methods are therefore increasingly used to improve speech quality, robustness, and controllability.

F5-TTS~\cite{chen2025f5} and E2 TTS~\cite{Eskimez2024E2TTS} simplify zero-shot TTS with flow-matching-based generation, reducing reliance on complex handcrafted pipelines. Voicebox~\cite{le2023voicebox} formulates speech generation as text-guided speech infilling and supports multilingual speech generation, editing, and denoising. NaturalSpeech 2~\cite{Shen2023NaturalSpeech2} and NaturalSpeech 3~\cite{Ju2024NaturalSpeech3} use latent diffusion and factorized codec representations to improve naturalness and zero-shot synthesis quality. DiTTO-TTS~\cite{Lee2024DiTToTTS}, Mask-GCT~\cite{wang2024maskgct}, SimpleSpeech 2~\cite{yang2025simplespeech2}, FlashSpeech~\cite{ye2024flashspeech}, DEX-TTS~\cite{Park2024DEXTTS}, and ARDiT~\cite{liu2024autoregressive} further explore diffusion transformers, masked generation, scalar latent diffusion, consistency acceleration, style disentanglement, and autoregressive diffusion. In video translation, these rendering techniques are valuable because speech must be not only natural, but also stable under duration constraints and suitable for downstream lip synchronization.

\begin{figure*}[t!]
  \centering
  \includegraphics[width=.82\textwidth]{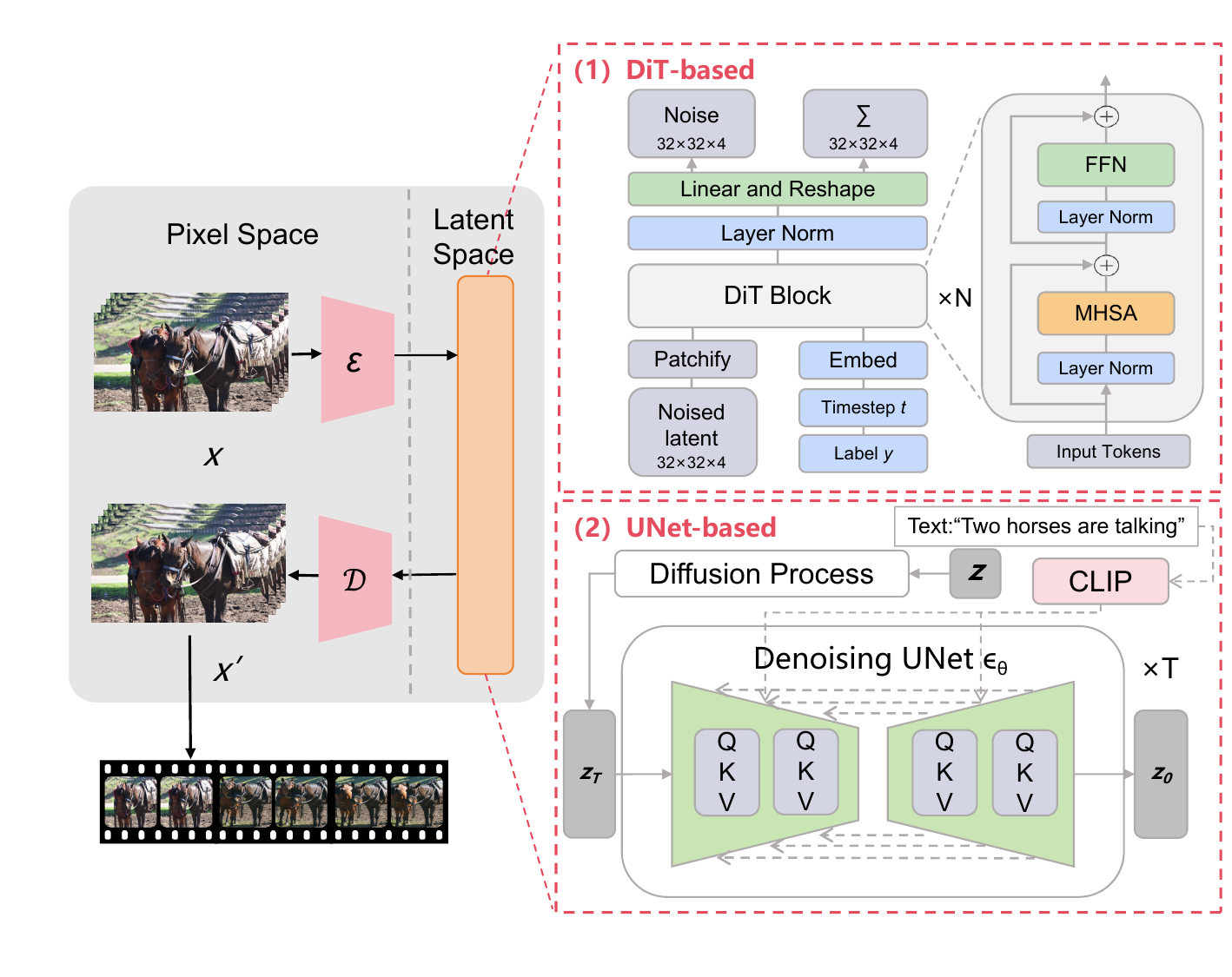}
  \caption{Comparison between UNet-based and DiT-based enabling visual backbones. UNet-based models commonly employ convolutional encoder--decoder denoisers with temporal modules, whereas DiT-based models represent videos as spatio-temporal latent tokens and perform denoising with Transformer blocks. In video translation, these backbones are mainly relevant when adapted for audio-driven lip motion, identity-preserving talking-head animation, and temporally coherent speaker rendering.}
  \label{fig:visual_backbones}
\end{figure*}

\subsection{The Visual Synthesizer}
\label{sec:visual_synthesizer}

The Visual Synthesizer is responsible for converting translated speech and multimodal control signals into visually coherent speaker motion. Unlike general video generation, visual synthesis in video translation is constrained by the source video, the target-language speech, and the identity of the original speaker. A high-quality system must align mouth movements with translated phonemes, preserve facial identity and expression, maintain temporal consistency across frames, and handle duration mismatches introduced by cross-lingual translation. Therefore, this section focuses on three functionally relevant directions: lip synchronization, talking-head animation, and enabling visual backbones.

\subsubsection{Lip Synchronization}

Lip synchronization is the most direct visual requirement in video translation. Given the translated speech and the source video, lip-sync methods modify the mouth region so that visible articulatory motion matches the target-language audio. Early systems relied on phoneme-to-viseme rules or handcrafted facial animation, but they often failed under large pose variations, expressive speech, or cross-lingual duration changes. Recent neural methods learn audio-visual correspondences from data and generate more realistic mouth movements.

Representative methods include Wav2Lip~\cite{prajwal2020lip}, which uses a lip-sync expert to improve audio-visual synchronization in unconstrained videos, and VideoReTalking~\cite{cheng2022videoretalking}, which edits talking-head videos by combining expression editing and lip synchronization. TalkLip~\cite{zhou2023seeing} further introduces lip-reading guidance to strengthen the consistency between generated lip motion and spoken content. Diffusion-based methods such as Diff2Lip~\cite{mukhopadhyay2024diff2lip}, MuseTalk~\cite{zhang2024musetalk}, LatentSync~\cite{li2024latentsync}, and OmniSync~\cite{peng2025omnisync} improve visual quality by performing audio-conditioned generation or inpainting in latent spaces. For video translation, these methods are especially important because translated speech often differs from the source speech in phoneme structure, word order, and duration, making direct reuse of the original lip motion insufficient.

\subsubsection{Talking-head Animation}

While lip synchronization focuses mainly on mouth motion, talking-head animation aims to generate a more complete facial performance, including head pose, expression, eye movement, and identity-preserving facial dynamics. This is important for translated videos in which the target speech must not only match the lips, but also remain consistent with the speaker's emotion, style, and visual identity.

Few-shot talking-head generation~\cite{zakharov2019few} shows that realistic neural talking heads can be synthesized from limited identity-specific data. IP-LAP~\cite{nie2023identity} improves identity preservation by incorporating landmark and appearance priors. More recent avatar-oriented systems, such as Kling-Avatar~\cite{ding2025kling}, OmniHuman-1~\cite{lin2025omnihuman1}, MIDAS~\cite{chen2025midas}, and Empathyear~\cite{fei2024empathyear}, extend facial animation toward long-duration, instruction-driven, or interactive digital-human generation. In the context of video translation, these systems provide a path beyond mouth-only editing: they can potentially coordinate translated speech with facial expression, head motion, and communicative intent, which is crucial for natural dubbed videos and multilingual virtual avatars.

\subsubsection{Enabling Visual Backbones}

General video generative models are not video translation systems by themselves, but their backbone designs provide important mechanisms for high-fidelity and temporally coherent visual synthesis. As shown in Fig.~\ref{fig:visual_backbones}, two representative families are UNet-based diffusion models and diffusion Transformer (DiT)-based models. The backbone discussion builds on UNet-based image generation, latent diffusion, diffusion Transformers, and early text-to-image or text-to-video generation studies~\cite{ronneberger2015u,rombach2022high,peebles2023scalable,ramesh2021zero,yu2022scaling,singer2022make,esser2023structure,hong2022cogvideo}.

UNet-based video diffusion models extend image diffusion architectures with temporal layers, motion modules, or latent-space video representations. Models such as Stable Video Diffusion~\cite{blattmann2023stable}, AnimateDiff~\cite{guo2023animatediff}, VideoCrafter~\cite{chen2023videocrafter1}, and Tune-A-Video~\cite{wu2023tune} demonstrate that pretrained image generators can be adapted to video generation while preserving spatial fidelity. Their relevance to video translation lies in localized editing, identity-preserving inpainting, and temporally stable facial rendering.

DiT-based video generation models replace convolutional denoisers with Transformer-based denoising over spatio-temporal latent tokens. Representative models such as CogVideoX~\cite{yang2024cogvideox}, Vidu~\cite{bao2024vidu}, Phantom~\cite{liu2025phantom}, HunyuanVideo~\cite{kong2024hunyuanvideo}, Wan~\cite{wan2025wan}, and Prompt-A-Video~\cite{ji2024promptavideo} improve long-range temporal modeling and text-video alignment. For video translation, these properties are useful when visual synthesis must maintain consistent speaker identity, facial motion, and scene-level continuity across longer translated segments. However, because most of these models are designed for general video generation rather than audio-driven translated-video editing, their direct application still requires stronger audio conditioning, identity constraints, and controllable synchronization mechanisms.

\section{Benchmarks, Metrics, and Evaluation Protocols}
\label{sec:evaluation}

Evaluating MLLM-enabled video translation is challenging because the task involves semantic transfer, expressive speech generation, and visual synchronization at the same time. Existing studies rarely provide a unified end-to-end benchmark for translated videos. Therefore, current evaluation is mostly role-level and proxy-based: video question answering benchmarks are used to assess semantic reasoning, speech datasets and perceptual metrics are used to evaluate expressive performance, and lip-sync or video generation metrics are used to measure visual synthesis quality. This section summarizes these evaluation practices according to the three roles in our taxonomy and discusses their limitations for end-to-end video translation tasks.

\begin{table*}[t]
\centering
\caption{Zero-shot open-ended VideoQA accuracy of representative MLLMs. Higher scores are better.}
\label{tab:qa_evaluation}
\renewcommand{\arraystretch}{1.08}
\setlength{\tabcolsep}{5pt}
\begin{threeparttable}
\begin{tabular}{@{}lcccc@{}}
\toprule
\textbf{Model} &
\textbf{MSVD-QA~\cite{xu2017msvdqa}} &
\textbf{MSRVTT-QA~\cite{xu2017msvdqa}} &
\textbf{ActivityNet-QA~\cite{yu2019activitynet}} &
\textbf{Avg.} \\
\midrule
\multicolumn{5}{@{}l}{\emph{Audio-visual and omni-modal MLLMs}} \\
FAVOR~\cite{sun2023fine} & 67.8 & 59.3 & -- & 63.5 \\
Dolphin~\cite{guo2025aligned} & 72.7 & 62.6 & 49.1 & 61.5 \\
Video-SALMONN~\cite{sun2024video} & 67.9 & 59.5 & -- & 63.7 \\
AV-LLM~\cite{shu2025audio} & 67.3 & 53.7 & 47.2 & 56.1 \\
OneLLM~\cite{han2024onellm} & 56.5 & 53.8 & -- & 55.1 \\
\midrule
\multicolumn{5}{@{}l}{\emph{Video-language alignment and grounding}} \\
Video-ChatGPT~\cite{maaz2024video} & 64.9 & 49.3 & 35.2 & 49.8 \\
Video-LLaMA~\cite{zhang2023video} & 51.6 & 29.6 & 12.4 & 31.2 \\
LLaMA-VID~\cite{li2024llama} & 69.7 & 57.7 & 47.4 & 58.3 \\
VideoChat2~\cite{li2024mvbench} & 70.0 & 54.1 & 49.1 & 57.7 \\
IG-VLM~\cite{kim2024image} & \textbf{76.7} & 62.7 & 57.3 & \textbf{65.6} \\
PLLaVA~\cite{xu2024pllava} & \underline{76.6} & 62.0 & 56.3 & \underline{65.0} \\
\midrule
\multicolumn{5}{@{}l}{\emph{Temporal and event reasoning}} \\
MovieChat~\cite{song2024moviechat} & 61.0 & 49.7 & 51.5 & 54.1 \\
LongVLM~\cite{weng2024longvlm} & 70.0 & 59.8 & 47.6 & 59.1 \\
VTimeLLM~\cite{huang2024vtimellm} & 69.8 & 58.8 & 45.5 & 58.0 \\
ST-LLM~\cite{liu2024st} & 74.6 & 63.2 & 50.9 & 62.9 \\
Slot-VLM~\cite{xu2024slot} & 74.9 & \textbf{69.7} & 48.3 & 64.3 \\
VideoLLaMA 3~\cite{zhang2025videollama3frontiermultimodal} & -- & -- & \underline{58.2} & 58.2 \\
GPT4-Video~\cite{wang2024gpt4video} & -- & -- & \textbf{59.5} & 59.5 \\
\midrule
\multicolumn{5}{@{}l}{\emph{Efficient adaptation and other representative models}} \\
LLaMA-Adapter~\cite{zhang2023llama} & 54.9 & 43.8 & 34.2 & 44.3 \\
PG-Video-LLaVA~\cite{munasinghe2023pg} & 64.1 & 51.6 & 39.9 & 51.9 \\
RED-VILLM~\cite{huang2024image} & 71.2 & 53.9 & 44.2 & 56.4 \\
MiniGPT4-video~\cite{ataallah2024minigpt4} & 73.9 & 58.8 & 44.4 & 59.0 \\
\bottomrule
\end{tabular}

\end{threeparttable}
\end{table*}

\subsection{Semantic Reasoner Evaluation}

The Semantic Reasoner is usually evaluated through video understanding and video question answering benchmarks. These benchmarks measure whether a model can identify objects, actions, events, temporal relations, and scene-level context from video input. Although they are not designed specifically for video translation, they provide useful proxy evidence for evaluating whether a model can ground translation decisions in visual and temporal information.

Representative benchmarks include MSRVTT-QA~\cite{xu2017msvdqa}, MSVD-QA~\cite{xu2017msvdqa}, and ActivityNet-QA~\cite{yu2019activitynet}. MSRVTT-QA is built on MSR-VTT~\cite{Xu2016MSRVTTPaper} and contains diverse open-domain videos with large-scale question--answer pairs, making it suitable for evaluating general video-language understanding. MSVD-QA is derived from MSVD~\cite{Chen2011MSVD} and focuses on short, relatively simple clips, which makes it useful for testing basic visual grounding. ActivityNet-QA contains longer untrimmed videos and questions requiring temporal and event-level reasoning, making it more relevant to realistic video translation scenarios where speaker actions, scene transitions, and temporal dependencies affect translation choices.

However, high performance on VideoQA benchmarks does not necessarily imply high-quality video translation. These benchmarks mainly evaluate visual-language understanding, while video translation additionally requires cross-lingual adequacy, timing control, speaker consistency, and audiovisual alignment. Therefore, the results in Table~\ref{tab:qa_evaluation} should be interpreted as evidence of semantic reasoning ability rather than direct evidence of end-to-end translation quality.

\subsection{Expressive Performer Evaluation}

The Expressive Performer is evaluated by the quality, intelligibility, speaker similarity, and controllability of synthesized speech. Unlike ordinary TTS evaluation, video translation requires the generated speech to preserve not only linguistic content, but also speaking rhythm, emotion, speaker identity, and duration constraints imposed by the source video.
Table~\ref{tab:performer_models} summarizes representative Expressive Performer models according to their functional roles in video translation. Speech-token language models mainly improve zero-shot voice cloning and scalable speech generation; prompt- and emotion-control models provide explicit control over speaking style and affect; diffusion- and flow-based renderers improve acoustic fidelity and stability. These capabilities are complementary for video translation, where the generated speech must preserve target-language meaning while matching speaker identity, rhythm, emotion, and video timing constraints.

\begin{table*}[t]
\centering
\caption{Representative Expressive Performer models for speech synthesis and controllable speech generation.}
\label{tab:performer_models}
\footnotesize
\renewcommand{\arraystretch}{1.12}
\setlength{\tabcolsep}{4pt}
\begin{tabularx}{\linewidth}{@{}p{2.7cm}p{2.4cm}cp{3.2cm}X@{}}
\toprule
\textbf{Method} & \textbf{Category} & \textbf{Zero-shot} & \textbf{Main Control} & \textbf{Key Architecture} \\
\midrule
\multicolumn{5}{@{}l}{\emph{Speech-token Language Models}} \\
CosyVoice~\cite{FunAudioLLMCosyVoice} & Speech-token LM & $\checkmark$ & Speaker, multilingual style & Semantic tokenizer + text-to-token LLM + flow matching decoder \\
CosyVoice 2~\cite{Du2024CosyVoice2} & Speech-token LM & $\checkmark$ & Streaming, speaker style & FSQ tokenizer + streamlined LLM + chunk-aware flow matching \\
Spark-TTS~\cite{Wang2025SparkTTS} & Speech-token LM & $\checkmark$ & Speaker, acoustic attributes & Qwen-based LLM + BiCodec tokenizers \\
VALL-E R~\cite{han2024vall} & Speech-token LM & $\checkmark$ & Speaker prompt & Neural codec LM + monotonic alignment \\
HALL-E~\cite{Nishimura2024HALLE} & Speech-token LM & $\checkmark$ & Long-form speech & Hierarchical neural codec LM \\
\midrule
\multicolumn{5}{@{}l}{\emph{Prompt and Emotion Control}} \\
PromptTTS~\cite{Guo2022PromptTTS} & Prompt control & $\times$ & Style prompt & BERT-based style encoder + speech decoder \\
PromptTTS 2~\cite{leng2023prompttts} & Prompt control & $\times$ & Descriptive prompt & NaturalSpeech 2 backbone + variation network \\
InstructTTS~\cite{Yang2023InstructTTS} & Prompt control & $\times$ & Natural-language style & Style encoder + discrete diffusion model \\
EMO-DPO~\cite{Gao2024EmoDPO} & Emotion control & $\times$ & Emotion preference & Emotion-aware LLM-TTS + DPO fine-tuning \\
StyleTTS 2~\cite{Li2023StyleTTS2} & Style control & $\checkmark$ & Speaker, style & Style diffusion + adversarial speech LM discriminator \\
\midrule
\multicolumn{5}{@{}l}{\emph{Diffusion and Flow Rendering}} \\
F5-TTS~\cite{chen2025f5} & Flow rendering & $\checkmark$ & Speaker prompt & ConvNeXt encoder + flow matching DiT \\
E2 TTS~\cite{Eskimez2024E2TTS} & Flow rendering & $\checkmark$ & Speaker prompt & Filler-token modeling + flow matching generator \\
Voicebox~\cite{le2023voicebox} & Flow rendering & $\checkmark$ & Speech infilling & Flow matching for text-guided speech infilling \\
NaturalSpeech 2~\cite{Shen2023NaturalSpeech2} & Diffusion rendering & $\checkmark$ & Speaker prompt & Codec encoder-decoder + latent diffusion \\
Mask-GCT~\cite{wang2024maskgct} & Codec generation & $\checkmark$ & Speaker prompt & Masked generative codec Transformer \\
\bottomrule
\end{tabularx}
\end{table*}

\begin{table}[t]
\centering
\caption{Representative speech datasets for Expressive Performer evaluation.}
\label{tab:performer_datasets}
\renewcommand{\arraystretch}{1.12}
\setlength{\tabcolsep}{3.5pt}
\begin{tabularx}{\columnwidth}{@{}p{0.27\columnwidth}
                                p{0.19\columnwidth}
                                p{0.16\columnwidth}
                                >{\raggedright\arraybackslash}X@{}}
\toprule
\textbf{Dataset} & \textbf{Language} & \textbf{Scale} & \textbf{Primary Use} \\
\midrule
LibriLight~\cite{kahn2020libri} & English & 60k h & Speech Pretraining \\
LibriTTS~\cite{Zen2019LibriTTS} & English & 585 h & Speaker Synthesis \\
LibriSpeech~\cite{Panayotov2015LibriSpeech} & English & 1k h & ASR and Intelligibility \\
MLS~\cite{pratap2020mls} & Multilingual & Large-scale & Multilingual Speech \\
Emilia~\cite{he2024emilia} & Multilingual & Large-scale & Zero-shot TTS \\
\bottomrule
\end{tabularx}
\end{table}
Training and evaluation datasets for this role usually come from speech recognition, speech synthesis, and speech representation learning. LibriLight~\cite{kahn2020libri} provides large-scale unlabeled speech for self-supervised acoustic representation learning. LibriSpeech~\cite{Panayotov2015LibriSpeech} is widely used for ASR and intelligibility evaluation. LibriTTS~\cite{Zen2019LibriTTS}, derived from LibriSpeech with improved segmentation and punctuation, is commonly used for TTS training and evaluation. These datasets support robust speech modeling, but they are mostly audiobook-style English corpora and do not fully cover multilingual, conversational, emotional, noisy, or video-synchronized speech conditions.

Common evaluation metrics include mean opinion score (MOS) for naturalness, word error rate (WER) for intelligibility, speaker similarity for identity preservation, and prosodic measures such as pitch and duration consistency. For controllable TTS, emotion accuracy and style similarity are also important. Nevertheless, zero-shot results are difficult to compare across papers unless the same speakers, prompts, languages, and evaluation protocols are used. For video translation, future benchmarks should evaluate whether synthesized speech matches the translated semantics, the source speaker identity, and the temporal structure of the original video.

\subsection{Visual Synthesizer Evaluation}

The Visual Synthesizer is evaluated by how well the translated speech is visually rendered on the speaker's face and body. Existing evaluation usually focuses on lip synchronization, visual fidelity, identity preservation, and temporal consistency. Lip-sync methods are commonly assessed with SyncNet-based metrics such as LSE-D and LSE-C, which estimate the alignment between audio and mouth motion. Talking-head and video generation methods are often evaluated with image or video quality metrics such as FID, FVD, identity similarity, and perceptual similarity.

These metrics are useful but incomplete. A model may obtain strong lip-sync scores while still producing unnatural expressions, identity drift, temporal flicker, or mismatched emotion. Conversely, a visually realistic talking-head model may fail to align precisely with target-language phonemes. Therefore, visual synthesis evaluation for video translation should jointly consider audio-visual synchronization, identity preservation, expression consistency, and robustness to cross-lingual duration mismatch.

\vspace{-3mm}

\subsection{End-to-End Evaluation Gaps}

Current role-level benchmarks provide useful diagnostic signals, but they do not fully measure the quality of translated videos. A complete evaluation protocol should combine automatic metrics with human assessment across multiple dimensions: semantic adequacy, translation fluency, speech naturalness, speaker similarity, emotional consistency, lip synchronization, visual fidelity, temporal coherence, and overall user preference.

Overall, existing evaluation remains fragmented. VideoQA benchmarks do not test translation quality, speech metrics do not evaluate visual grounding, and lip-sync metrics do not measure semantic adequacy. A major open direction is therefore to develop end-to-end multilingual video translation benchmarks with aligned source videos, target speech, visual outputs, and human evaluation protocols.

\section{Limitations and Future Directions}
\label{sec:limitations}

Although MLLMs provide a promising foundation for video translation, current systems remain far from reliable, natural, and deployable cross-lingual video communication. Existing methods are still limited by incomplete video understanding, weak temporal and multimodal alignment, fragmented evaluation protocols, insufficient multilingual robustness, high deployment cost, and growing safety concerns. This section discusses these limitations and outlines future directions.

\subsection{Fine-grained and Long-form Video Understanding}

Accurate video translation requires models to understand not only spoken content, but also visual context, speaker behavior, emotion, gestures, object interactions, and scene transitions. Current MLLMs still struggle with fine-grained visual semantics~\cite{tang2025caption,qian2024momentor,sun2023fine} and long-form video understanding~\cite{song2024moviechat,weng2024longvlm,song2024moviellm,ren2024timechat,qian2024streaming,ma2025drvideo}. These limitations are particularly harmful for translation because ambiguous utterances, deixis, speaker turns, and visually grounded references often cannot be resolved from ASR transcripts alone.

Future work should develop hierarchical and event-aware video representations that combine global narrative structure with local semantic details. Adaptive frame sampling~\cite{han2024self}, attention-guided token selection~\cite{huang2024ivtp}, memory-augmented modeling, and retrieval-based context expansion may reduce redundant visual tokens while preserving translation-relevant information. More importantly, future datasets should provide temporally grounded annotations of speakers, events, emotions, and visual references, enabling models to learn which visual cues should influence translation decisions.

\subsection{Temporal and Cross-modal Alignment}

Temporal alignment is central to video translation. A translated sentence must remain synchronized with source speech, speaker motion, facial expression, and visual events. However, current systems often model video frames, audio signals, ASR transcripts, and translated text as loosely connected streams. This makes it difficult to handle rapid scene changes, overlapping speech, long-range dependencies, and cross-lingual duration mismatch. Existing temporal modeling methods use frame indices, timestamp prompts, relative positional encoding~\cite{hao2024posmlp}, or sequence-context learning~\cite{lin2024learning}, but there is still no unified representation for linking linguistic units with audiovisual events.

Future research should move toward explicit audio-visual-text alignment models that represent duration, order, rhythm, speaker turns, and causal event structure. Such models should support both local alignment, such as phoneme-to-viseme synchronization, and global alignment, such as maintaining coherent translation across long scenes. Cross-modal attention, event-level abstraction~\cite{wang2024modeling}, and streaming temporal memory may help MLLMs align translated outputs with dynamic video context more precisely.

\subsection{End-to-end Evaluation and Benchmarks}

A major bottleneck is the lack of end-to-end evaluation protocols for MLLM-enabled video translation. Current studies typically rely on proxy benchmarks: VideoQA for semantic reasoning, ASR or MT metrics for linguistic quality, MOS for speech naturalness, and SyncNet-based scores for lip synchronization. However, strong performance on these isolated metrics does not guarantee high-quality translated videos. For example, a model may achieve high lip-sync scores while producing unnatural expressions, or strong VideoQA results while failing to preserve cross-lingual meaning.

Future benchmarks should evaluate translated videos as complete multimodal artifacts. A desirable protocol should jointly measure semantic adequacy, translation fluency, speaker similarity, emotional consistency, lip synchronization, visual fidelity, temporal coherence, and user preference. Human evaluation remains necessary, but it should be standardized with clear rubrics and multilingual test sets. Building public end-to-end benchmarks with source videos, target translations, dubbed speech, visual outputs, and human judgments would substantially improve comparability across systems.

\subsection{Multilingual, Low-resource, and Cultural Generalization}

Most existing video translation research is biased toward high-resource languages, clean speech, and relatively standard visual scenarios. Real-world videos often contain accents, dialects, code-switching, cultural references, humor, politeness strategies, gestures, and domain-specific expressions. These factors make video translation more complex than literal cross-lingual conversion. A system that is semantically correct at the sentence level may still fail pragmatically if it ignores cultural context or mismatches the speaker's communicative intent.

Future work should expand beyond high-resource language pairs and evaluate low-resource, accented, multilingual, and code-switched scenarios. Culturally aware translation should incorporate visual grounding, discourse context, and pragmatic intent, especially for humor, emotion, honorifics, and gesture-related meaning. Cross-lingual transfer, synthetic data generation, and human-in-the-loop refinement may help reduce annotation cost while improving robustness across languages and cultural settings.

\subsection{Real-time and Resource-efficient Deployment}

Many video translation applications, such as livestreaming, online meetings, education, and customer service, require low latency and stable streaming performance. However, MLLM-enabled systems are computationally expensive because they must process long video sequences, audio streams, transcripts, translation history, and generation modules simultaneously. High-resolution video understanding and diffusion-based visual synthesis further increase memory and inference cost~\cite{wu2024videollm}.

Future systems should be designed for streaming and resource-constrained deployment from the beginning. Promising directions include structured sparsity~\cite{Zheng2024StructuredSparsity}, adaptive token pruning, lightweight temporal memory, incremental decoding, and streaming speech translation~\cite{Tripathi2020TransformerTransducer}. Instead of processing entire videos offline, models should selectively update translation-relevant context and generate speech or visual edits with controllable latency. This direction is essential for moving MLLM-enabled video translation from offline demos to practical applications.

\subsection{Responsible and Trustworthy Video Translation}

Video translation technologies can alter a speaker's voice, face, language, and perceived intent, which creates serious ethical and safety risks. Voice cloning, lip synchronization, and talking-head generation may be misused for impersonation, misinformation, or unauthorized identity manipulation. Even benign translation systems may distort speaker identity, emotion, or cultural meaning if consent, attribution, and transparency are not properly handled.

Future research should incorporate responsible design principles into the technical pipeline. Important directions include consent-aware voice and face cloning, watermarking for synthesized audio and video, provenance tracking, misuse detection, identity protection, and user-controllable translation styles. Evaluation should also include trustworthiness dimensions, such as whether the translated video preserves the speaker's intent and whether synthetic modifications are clearly disclosed. 

\section{Conclusion}
\label{sec:conclusion}
This survey reviewed how MLLMs can reshape video translation from a cascaded processing pipeline into a role-oriented multimodal reasoning and generation framework. We organized existing studies around three functional roles: Semantic Reasoner, Expressive Performer, and Visual Synthesizer, and analyzed how they contribute to semantic fidelity, temporal alignment, speech expressiveness, and visual synchronization. We also summarized representative datasets, benchmarks, and metrics, highlighting the gap between current proxy evaluations and end-to-end video translation quality. Finally, we discussed open challenges in fine-grained and long-form understanding, cross-modal alignment, end-to-end evaluation, multilingual generalization, real-time deployment, and responsible use. We hope this survey provides a coherent conceptual framework and practical roadmap for developing natural, reliable, and trustworthy cross-lingual video translation systems.

\bibliographystyle{IEEEtran}
\bibliography{refs}

\end{document}